\documentclass[11pt]{article}
\usepackage{coling2018}
\usepackage{times}
\usepackage{url}
\usepackage{latexsym}
\usepackage{graphicx, dblfloatfix}
\usepackage{algorithm}
\usepackage{algorithmic}
\usepackage{booktabs} 

\title{A Dialogue Annotation Scheme for Weight Management Chat using the Trans-Theoretical Model of Health Behavior Change}

\author{Ramesh Manuvinakurike$^{1*}$, Sumanth Bharadwaj$^2*$, Kallirroi Georgila$^1$ \\
$^1$Institute for Creative Technologies, University of Southern California \\
  {\tt $^1$[manuvinakurike,kgeorgila]@ict.usc.edu, $^2$first\_last@yahoo.com} \\}
  
\date{}

\begin{document}
\maketitle
\begin{abstract}
  In this study we collect and annotate human-human role-play dialogues in the domain of weight management. 
  There are two roles in the conversation: the ``seeker'' who is looking for ways to lose weight and the ``helper'' who provides suggestions to help the ``seeker'' in their weight loss journey.
  The chat dialogues collected are then annotated with a novel annotation scheme inspired by a popular health behavior change theory called ``trans-theoretical model of health behavior change''. 
  We also build classifiers to automatically predict the annotation labels used in our corpus. We find that classification accuracy improves when oracle segmentations of the interlocutors' sentences are provided compared to directly classifying unsegmented sentences. 
\end{abstract}

\section{Introduction}
\label{sec:intro}
\blfootnote{$^*$ Equal contribution.}
Individuals seeking ways to modify their unhealthy lifestyles are interested in the personal experiences of other people who describe how they have changed their unfavorable health behavior (e.g., smoking, poor diet, overeating, etc.).
Such experiences are shared as stories where a person who successfully changed their health behavior narrates the journey from an unfavorable to a more favorable lifestyle in a blog or posts in a public forum.
There are thousands of such stories. 
But different stories may have a different impact depending on who reads them. Not every story is relevant to an individual, but rather only a few stories can successfully motivate and provide useful information to a specific reader.
This is because people tend to be influenced more by stories related to their personal experiences \cite{manuvinakurike2014automated}.
Research has shown that such personalized stories delivered to individuals are effective in motivating people to change their unfavorable health behavior successfully \cite{houston2011culturally}. 

People subscribe to the personal experiences of others and seek to gain motivation to change their unfavorable health behavior to an alternative favorable behavior. 
They do this by looking for the ``right'' processes that they can benefit from and include in their own lives, e.g., the following advice is suitable for a regular coffee drinker rather than someone who does not drink coffee:  ``drinking coffee in smaller portions helped me lose weight''. 
A recent survey also showed that individuals trust the stories and experiences shared on the ``internet'' (by professional health advisers) more than the stories and experiences of a family member or friend \cite{fox2011social}. 
This is not so surprising as the plethora of stories available on the internet makes it easier for users to find the ``right'' story that they can relate to. 

Such a phenomenon of sharing stories and experiences on the internet is typically observed in social health advice sharing forums where a user with an unfavorable health behavior submits a post describing their problem.
This is followed by peers replying to the post with their own relatable stories or providing suggestions based on their personal experiences. 
Such forums offer users a platform to share their stories, and provide help and encouragement to other users. 
Seeking help in this way could prove to be effective in motivating people to change to a healthier lifestyle such as weight loss \cite{hwang2010social}. 
These forums are popular and continue to gain in popularity. 
However, while these forums are useful, a real-time conversation where users can engage in a session of question answering and experience sharing could potentially yield more benefits. 

There is growing interest in building conversational agents that can motivate users to change their health behaviors; see for example \newcite{rizzo2011simcoach}. Such systems typically have expert-authored content.
However, generating expert-authored content tailored to an innumerable number of users is a difficult (if not impossible) task. 
For this reason there are benefits in pursuing the development of automated chat bots that can engage users in a conversation about changing their health behaviors, and that can be trained on chat-based conversations between regular people (non-experts) who exchange information in order to help one another.
It is important to note that we do not claim that such a chat bot could replace an expert but rather act as a companion that could assist users by providing useful pointers towards their goal.

Obesity continues to grow in epidemic proportions. 
A change in lifestyle could help the population, and this serves as a long-term motivation for our work.
We envision agents conversing with and assisting humans by providing advice, stories, and tips that the individuals can benefit from. 
Once developed, these agents can be used to (i) motivate users to begin their weight loss journey; (ii) provide useful tips on lifestyle changes to users already contemplating to lose weight; (iii) provide advanced suggestions and tips to users already in the process of losing weight; (iv) provide encouragement by reminding users about their goals; and (v) help users maintain their target weight once their goal is reached.
However, developing such agents is a challenging problem. 
The agents need to carefully diagnose the condition of the person seeking to change their health behavior. Based on a variety of constraints the agent has to offer relevant and beneficial advice. 
For instance, behavior change advice for weight loss to an individual who is obese and looking to lose weight will be very different from the tips offered to an individual who is looking to lose the last few pounds to reach a fat percentage of less than 10\%. 

Such a ``recommendation system'' can benefit from research in health behavior change psychology. 
Advising health behavior change has been studied in the literature of psychology, and various competing theories exist which act as guidebooks in providing the right advice to  individuals. 
One such theory, that we make use of in this work, is called ``trans-theoretical model of health behavior change'' \cite{prochaska1997transtheoretical}.
This theory has proven successful in modeling health behavior change in individuals and provides a model for such a process \cite{prochaska1997transtheoretical,johnson2008transtheoretical,tuah2011transtheoretical,mastellos2014transtheoretical}.
The theory also provides a mapping from the person's stage in the journey of health behavior change to the classes of actionable items that can be taken to progress from an unfavorable stage to a favorable stage. 
For example, if the person is contemplating a change in their eating habits but still has not fully committed to the change in their behavior, one of the better pieces of advice to offer to this person is raising the awareness of the ill effects of being on a poor diet. This type of advice would not be relevant to a person who is acting upon their diet plan and is fully aware of the ill effects of a poor diet. 

Research in the field of health behavior psychology has been carried out extensively by studying weight loss behaviors in humans. 
Recently the trans-theoretical model of health behavior change has been used to guide research on virtual human agents for studying the motivation of individuals with regard to weight management \cite{bickmore2005establishing,bickmore2006health,bickmore2013automated,manuvinakurike2014automated}. 
Our approach is different from the approaches followed in these works. In our work we annotate human-human chats based on the trans-theoretical model whereas previous work used concepts from the trans-theoretical model to guide agent utterances that were authored by human experts. 
In our work, we develop a corpus containing typed chats between a human health behavior change seeker and a helper. The chat is annotated using labels motivated from dialogue research \cite{bunt2012iso}
and the trans-theoretical model of health behavior change for weight management \cite{prochaska1997transtheoretical}.

Our contributions are as follows: (i) development of a corpus containing dialogues between peers in a weight loss advice seeking session set up as a role-play game between a help seeker and a help provider; (ii) a novel annotation scheme for annotating the corpus; and (iii) models for automatically classifying sentences to one of the annotation labels used in our corpus. 
The rest of the paper is organized as follows. We begin with describing the data collection methods and experiments in Section~\ref{sec:datacollection}. We then describe the trans-theoretical model of health behavior change and our novel annotation scheme inspired by this model in Section~\ref{sec:annotation}.
Section~\ref{sec:experiments} describes our classification experiments.
Finally, we conclude and outline our plans for future work in Section~\ref{sec:future}. 

\section{Data Collection}
\label{sec:datacollection}

By collecting chat data we can model not only phenomena based on the trans-theoretical model of health behavior change but also conversational phenomena (e.g., question answering, acknowledgments, etc.) which are usually absent in data from  popular social media forum posts. 
In this work, we use a role-play scenario in a make-belief setting to collect our chat-based dialogue data. 

Crowd-sourcing has recently emerged as a popular platform for collecting dialogue data. 
It has also been popular among researchers studying the health behavior change phenomenon \cite{crequit2018mapping}. 
We collect data using the Amazon Mechanical Turk (MTurk) crowd-sourcing platform.
The task is set up as a role-play chat game between two turkers (users of MTurk). All the turkers got equal pay irrespective of the quality of their chat. 
The users on the MTurk are instructed that they will be either assigned the role of a ``help seeker'' (seeker) or ``help provider'' (helper). 
If they are assigned the role of the ``help seeker'', they are instructed to imagine a scenario where they are overweight and want to lose weight. They are required to have a conversation with their partner to seek help with their weight loss journey. 
They are also informed that they will be paired with a ``help provider'' who will assist them with tips to overcome their barriers and help them lose weight. 
If the users are assigned the role of a ``help provider'' they are instructed to play the role of a helper who needs to assist the ``help seeker'' with their goal of losing weight. 

Initially we were skeptical about the quality of chat that would result from such a setup. 
Surprisingly, the chat conversations between the participants yielded very good quality interactions indicating that MTurk could be a good platform for collecting similar chat data sets. The quality of the interaction was measured subjectively. 
The users were instructed to be kind to one another, and were informed that any abuse would disqualify them from participating in the experiment. 
The users were also asked to maintain anonymity and not reveal their name or personally identifiable information as the chat data could be released to the public in the future. 
The users were from the United States and native English speakers. 
Further demographic information about age and gender were not collected to maintain their anonymity. Once collected, the chat data were annotated by experts using the annotation scheme described in Section~\ref{sec:annotation}. Table~\ref{tab:stats} shows the statistics of the data collected and Table~\ref{tab:example} shows a snapshot of an example chat between a helper and a seeker. 

\begin{table}[t!]
    \centering
    \begin{tabular}{| l | c |}
        \hline
        \# users &  52 \\ \hline
        \# dialogues & 26 \\ \hline
        \# turns &  309 \\ \hline
        \# word tokens & 1230  \\ \hline
        average \# turns in a conversation & 10 \\ \hline
        \end{tabular}
        \caption{\label{tab:stats} Statistics of the corpus collected.}
\end{table}

\begin{table}[!ht]
    \centering
    \begin{tabular}{| l  p {12cm} |}
        \hline

        \textbf{role} & \textbf{chat} \\ \hline
        helper & [Greeting : Hello] \\
        seeker & [Greeting : Hello]! [Action : Just started my weight loss journey [TimeFrame : a couple months ago]] \\ 
        helper & [acknowledge : Thats fantastic!] [question : How is it going thus far?] \\ 
        ... &  \\ 
        seeker & [Contemplation : [goal : Id like to get down to 225]] \\ 
        seeker & [Contemplation : [SR : Ive been around [currWeight : 245-250] for [TimeFrame : years] now]] \\ 
        helper & [acknowledge : That makes sense!] [question : How much weight have you lost [TimeFrame : thus far?]] \\ 
        seeker & [Action : About 12 lbs] \\ 
        ... &  \\ 
        seeker & [Action : [question : How did you [SeLi : motivate yourself to work out?]]] \\ 
        helper & [Action : [SeLi : My motivation always came from changing things up]] \\ 
        helper & [Action : When [Lifestyle-undes : my music was no longer motivating], I found new music] \\ 
        helper &  [CC : [Lifestyle-undes : When I got bored of some exercises], I found new ones to try]  \\ 
        helper & [Action : [CC : When [Lifestyle-undes : I got sick of my [good-diet : diet]],  I found [good-diet : new foods]]] \\ 
        helper &  [Action : [SeLi : That always helped me to keep from feeling stuck]] \\ 
        seeker & [acknowledge : I can see how that would make a difference] \\ 
        seeker & [Preparation : [Lifestyle-undes : I tend to stick to one thing, but after [TimeFrame : a couple weeks], my motivation dies out]] \\ 
        helper & [question :  If your free time was at night before, what do you think about trying morning workouts for something new?] \\ 
        seeker &  [acknowledge : That could work]. [Lifestyle-undes : Im a little lethargic in the morning] \\ 
        seeker & [Contemplation : [Lifestyle-undes : Not exactly I morning person.] But I really dont have to think about getting on a treadmill - just have to do it] \\ 
        helper & [acknowledge : Nothing a little [good-diet : coffee] cant fix] :)  \\ 
        seeker & [End : Hey thank you for the motivation today!] \\ \hline
        \end{tabular}
        
        \caption{\label{tab:example} Example annotated chat interactions between a seeker and a helper exactly as they appear in the data set (with misspellings, lack of punctuation, etc.). SR: Self re-evaluation, SeLi: Self liberation, CC: Counterconditioning.}
\end{table}

\section{Annotation}
\label{sec:annotation}

Our novel annotation scheme was designed to leverage the benefits of the trans-theoretical model (TTM), which provides a theoretical framework for modeling the health behavior change process. The TTM also provides the framework for recommending activities to users based on their current stage in the journey of health behavior change. One of the goals of this annotation framework is to leverage the TTM's stages of change and processes of change. 

It is important to identify the seeker's current stage of change in order to offer theoretically motivated activity suggestions belonging to one of the processes of change also annotated in the data.
Likewise, it is also important to identify the processes for change recommended by the helper which form the activities that can be leveraged to motivate the seeker. 

Section~\ref{subsec:ttm} describes the TTM and its relation to our work. We use the stages of change and processes of change defined by the TTM as dialogue act labels. We also use traditional dialogue act labels such as questions, greetings, etc.\ to track the user state.

\subsection{Trans-Theoretical Model (TTM) of Health Behavior Change}
\label{subsec:ttm}

The two concepts of the TTM that we adopt in this work are called ``Stages Of Change'' (SOC) and ``Processes Of Change'' (POC). 
One of the requirements for performing this kind of annotation is familiarity with the TTM. The annotators need to study the TTM closely. This is one of the limitations of annotating a large data set based on the TTM.\footnote{Note that the SOC and POC can be identified deterministically by answering questions indicated in the TTM literature. This is something that we have not explored in our current work but keep in mind for our future work.}
The goal of the annotator is to correctly identify the seeker's ``stage of change'' (point in the journey of weight loss where the seeker is) based on the information that the seeker provides to the helper and the ``processes of change'' (i.e., activities recommended by the helper or indicated by the seeker).
Below we describe the SOC and POC labels used in this work. 

\subsubsection{Stages Of Change (SOC) }
The TTM places the users who want to change their health behavior into 5 stages, aptly called ``Stages Of Change'' (SOC). These stages are based on the individual's awareness and progress made during the health behavior change journey. They include changes from a stage where the individual is not aware that an unfavorable behavior needs to be changed to a stage where the change has been achieved and the individual is working towards avoiding a relapse back to the unfavorable health behavior. These 5 stages of change are:

\noindent
\textbf{i) Precontemplation}: People in this stage do not wish or do not know how to change their behavior. In our study the users are instructed specifically to role-play an individual who wants to change their behavior, and thus this stage is not observed in our data. 

\noindent
\textbf{ii) Contemplation}: In this stage the users are planning to change their behavior (typically within the next 6 months). Typically the user of such a ``health behavior change advice system'' is in at least this stage or further. 

\noindent
\textbf{iii) Preparation}: In this stage the users are taking action to change their behavior (typically in a month) and are susceptible to the majority of the processes of change (see Figure~\ref{fig:relation}). 

\noindent
\textbf{iv) Action}: In this stage the users have taken action to change their behavior and are making progress. They are no longer prone to advice about raising consciousness regarding the adverse effects of their unfavorable behavior. 

\noindent
\textbf{v) Maintenance}: In this stage the users have changed their behavior for at least 6 months and are working to avoid relapse. 

We identify the appropriate SOC based on the goals described by the seeker. Table~\ref{tab:example} shows an example where an individual states that the they have just started their ``weight loss journey a couple of months ago'' and hence places them in the ``Action'' SOC. Another example where the user says ``I have been wanting to change my behavior soon'' would place them in the ``Contemplation'' SOC. Such statements where users state their goals help the annotators place the seeker into one of the 5 SOC classes. 

\subsubsection{Processes Of Change (POC) }
The ``Processes Of Change'' (POC) refer to covert and overt activities that users engage in to progress through the SOC \cite{prochaska1997transtheoretical}. There are totally 10 processes of change that we use in this work:

\noindent
\textbf{1. Consciousness raising}: Attempt to seek out information concerning their problem behavior. Example: ``strength training is supposed to be great for getting in shape''.

\noindent
\textbf{2. Dramatic relief}: Increased emotional experiences followed by reduced affect if an appropriate action can be taken.
Examples: ``I'm worried about my health'', ``if I go to 250, I'm done with life''.

\noindent
\textbf{3. Substance use / Stimulus control}: Use of medication/devices/surgery (external substance). Removes cues for unhealthy habits and adds prompts for healthier alternatives.
Examples: ``I have found success with one of those items that count your steps everyday'', ``I'm thinking of trying a fitbit''.

\noindent
\textbf{4. Social liberation}: Increase in social opportunities.
Example: ``me losing weight will help my tag team perform well at a team event''. 

\noindent
\textbf{5. Self re-evaluation}: Cognitive and affective assessments of one's self-image. 
Example: ``I want to look like a shark''.

\noindent
\textbf{6. Helping relationships}: Combine caring, trust, openness, and acceptance as well as support for the healthy behavior change.
Examples: ``I'll have to find a partner'', ``yeah my mom does zumba and wants me to go''.

\noindent
\textbf{7. Counter conditioning}: Substituting an unfavorable health behavior with a favorable one. 
Example: ``juice has a lot of sugar but there are some different types of almond milks out there or even skim milk''.

\noindent
\textbf{8. Reinforcement management}: Consequences for taking steps in a particular direction.
Example: ``I do it by giving myself a cheat day only if I met my goals for the week''.

\noindent
\textbf{9. Self liberation}: The belief that one can change and the commitment and re-commitment to act on that belief.
Examples: ``believe it and have the dedication and you'll be able to succeed at it'', ``you have the ability to lose whatever you want''.

\noindent
\textbf{10. Environmental re-evaluation}: Affective and cognitive assessments of how the presence or absence of a personal habit affects one's social environment.
Example: ``my brother's band team were coming to visit him and I wanted to lose weight to make him look good''. 

\paragraph{Relation between SOC and POC: } 

  \begin{figure}[t!]
    \begin{center}
    \includegraphics[width=\columnwidth]{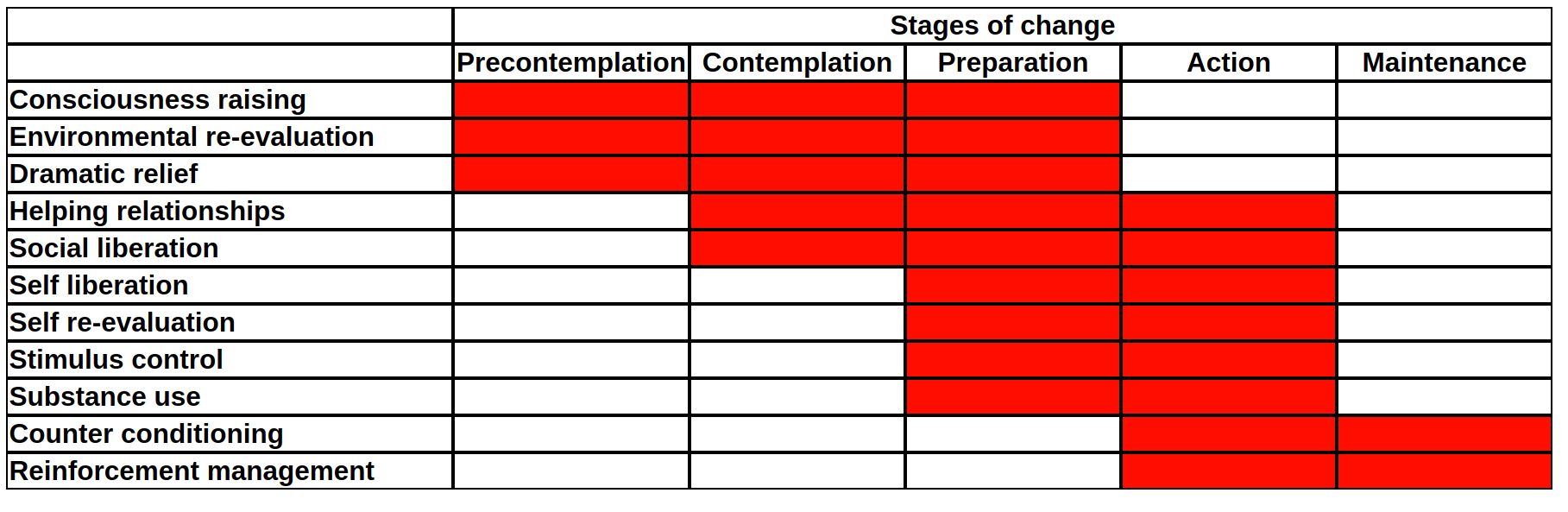}  
    \caption{Relation between the SOC and POC typically observed in users changing their health behavior. The red block indicates that a POC (row label) is commonly used in a given SOC (column label) for progression, whereas an empty white box indicates that the a POC (row label) is not commonly used in a given SOC (column label) for progression \cite{prochaska1997transtheoretical}.}
    \label{fig:relation}
    \end{center}
  \end{figure}  

Figure~\ref{fig:relation} shows the relation between the SOC and the POC in TTM\footnote{Figure~\ref{fig:relation} shows the ``Stimulus control'' POC that was not observed in our data.}. In this work we annotate the SOC of the seeker and the POC mentioned by both the helper and the seeker.
The POC annotations are designed to serve two purposes: (i) to equip future dialogue systems with the capability of providing suggestions based on the seeker's current SOC; and (ii) to track which POC were used by the seeker in the past or the seeker is aware of. 
The seekers in a given SOC are motivated to progress to the next stage by engaging in a POC. Generally, not all POC are suited for a given SOC. For instance, from Figure~\ref{fig:relation} we can observe that the POC ``consciousness raising'' is well suited for the individuals in the ``Precontemplation'', ``Contemplation'', and ``Preparation'' SOC. This is because further information about a behavior stops being useful for an individual in ``Action'' as the users in this SOC are already aware of the harmful effects of an unfavorable health behavior. This mapping between the SOC and POC is useful for identifying these labels in the data. However, it is important to keep in mind that the TTM provides the relations that we see in Figure~\ref{fig:relation} as a heuristic and not a rule to follow when performing the annotations. This implies that, while such a mapping is usually true, cases exist where a POC not indicated for a given SOC might be applicable. 

\subsection{``Other'' Labels}
\label{subsec:others}

We also identify ``other'' labels in the chat in order to facilitate better understanding of the seeker and helper behaviors. These labels are shown in Table~\ref{tab:others}. The table shows the labels, descriptions, and a relevant example. The labels include questions, greetings, end of the conversation markers, time information, etc. 

\begin{table}[!ht]
    \centering 
    \begin{tabular}{| l | p {5cm} | p{6.5cm} |}
        \hline
        \textbf{Label} & \textbf{Description} & \textbf{Example} \\ \hline
         
        question & question & how much weight are you looking to lose? \\ \hline 
        greeting & represents a greeting & how are you this evening? \\ \hline 
        goal & weight loss goals & I got back down to 190 \\ \hline
        time-frame & duration in time & a few months \\ \hline
	    bad-diet & bad dietary choices & sugar, fat \\ \hline
        good-diet & good dietary choices & vegetables \\ \hline
	    lifestyle-undesired & bad lifestyle choices & ate junk food \\ \hline
	    acknowledge & acknowledgments & yeah, I know \\ \hline
	    frequency & frequency of various behaviors & two days of the week \\ \hline
		end & 	end of conversation & thank you \\ \hline
	    device & equipment that aids weight loss & fitbit \\ \hline
	    current-weight & current weight & I'm 250 lbs \\
	    \hline
        \end{tabular}
        \caption{\label{tab:others} Additional ``other'' labels annotated in the data set.}
\end{table}

We measured the inter-annotator agreement using Cohen's kappa. It was found to be 0.66 for POC, 0.81 for SOC, and 0.72 for other labels annotated in the corpus. The values were calculated at the sentence level. The numbers showed good agreement between 2 expert annotators who were well versed at the TTM concepts and annotated the same 2 dialogues (22 turns). 

\paragraph{Shortcomings of the scheme:}
It was observed that a sentence could fall into multiple POC resulting in lower inter-annotator agreement, e.g., ``my mom is helping me eat broccoli for a snack instead of chips'',
falls under both ``helping relationships'' and ``counter conditioning''. 
Such cases caused disagreements between the annotators.
In order to account for this, further annotation labels would be needed or the annotation scheme would have to support annotation of each sentence with all applicable labels. However, these changes would make it hard to develop an automated classifier.

\section{Experiments}
\label{sec:experiments}

We performed machine learning experiments to automatically predict the annotation labels in our corpus. 
We build a separate classifier for SOC, POC, and ``other'' labels (3 classifiers in total). This is because these labels are annotated independent of one another.
Each classifier could output one of the corresponding labels or a ``null'' label. 
We use logistic regression in Weka \cite{hall2009weka},
and since no prior work exists a majority baseline for comparison. 
The data were preprocessed before the classification was performed. 
We used the NLTK toolkit for lemmatization \cite{Loper:2002:NNL:1118108.1118117} and removed stop words.
The features that we used were just words. We report the results on 10-fold cross validation performed on the user sentences. 

We predict the labels in two separate experiments: (i) ``unsegmented'' and (ii) ``segmented''. For both settings we use the same set of features. 
In the ``unsegmented'' version, we predict the classification labels using the complete sentences. Each complete sentence is forwarded to the 3 classifiers and each classifier outputs one of its corresponding labels or the ``null'' label.
For the ``segmented'' approach we segment the sentences and use each segment as an input to each classifier. Again each classifier outputs one of its corresponding labels or the ``null'' label.

In the ``segmented'' approach we assume oracle (perfect) segmentation of the user sentences before classification. In future experiments we plan to perform the segmentation automatically and then predict the label. 
Note however that the annotations can overlap, which means that an ``other'' label can be inside a section of the sentence annotated with a SOC or POC label. Similarly, POC and SOC labels can overlap. Hence we use 3 types of segmentations the output of which would be forwarded to each one of the 3 classifiers. Let us consider an example:

\noindent
[GREET\{OTHER\}: Hi there] [ACTION\{SOC\}: I would like to [GOAL\{OTHER\}: lose weight] but [SL\{POC\}: exercising] didn't help me much]

The segmentation for the SOC classifier would be: \\
\noindent
SEG1: Hi there\\
SEG2: I would like to lose weight but exercising didn't help me much

The segmentation for the POC classifier would be:\\
\noindent
SEG1: Hi there I would like to lose weight but\\
SEG2: exercising\\
SEG3: didn't help me much

The segmentation for the ``other'' label classifier would be:\\
\noindent
SEG1: Hi there \\
SEG2: I would like to\\
SEG3: lose weight\\
SEG4: but exercising didn't help me much

Table~\ref{tab:results} shows our results for each classifier:
``unsegmented'' majority baseline and accuracy using the ``segmented'' and ``unsegmented'' approaches. 
We observe that the ``segmented'' approach results in higher classification accuracies.

\begin{table}[t!]
    \centering
    \begin{tabular}{| l | c | c | c |}
        \hline
         \textbf{Task} & \textbf{Unsegmented} &\textbf{Unsegmented} &\textbf{Segmented} \\ 
        \textbf{} & \textbf{Majority} &\textbf{Classification} &\textbf{Classification} \\ 
        \textbf{} & \textbf{Baseline} &\textbf{Accuracy} &\textbf{Accuracy} \\ \hline
        SOC prediction & 0.37 & 0.44 & 0.48\\ \hline
        POC prediction & 0.25 & 0.41 & 0.49\\ \hline
        Other label prediction & 0.18 & 0.35 & 0.67 \\ \hline
        \end{tabular}
        \caption{\label{tab:results} Classification results. The differences between the unsegmented and the segmented accuracies as well as the differences between the unsegmented and segmented accuracies and the majority baseline are significant (p $<$ .05). }
\end{table}

\section{Conclusion \& Future Work}
\label{sec:future}

In this work we presented a novel annotation scheme for health behavior change motivation chat-based dialogues. Our annotation labels are grounded in the health behavior change psychology literature and are also complemented by standard annotation labels used in conversational data.
We also performed automated classification experiments using 3 classifiers for classifying SOC, POC, and ``other'' labels respectively.

We hypothesize that the sparsity of our data negatively impacts classification accuracy. In a follow up experiment we aim to expand our data set by collecting more chat interactions.
However, collection of large data sets can be an issue as our annotations require expert annotators. 
It will be fruitful to explore the possibility of extracting the annotation labels using crowd-sourcing by providing MTurk annotators with TTM-based questionnaires to guide their annotations. 
We also plan to extend this work by building a dialogue system that can play the role of the helper. 

\section*{Acknowledgments}
This work was partially
supported by the U.S. Army; statements and opinions expressed do not necessarily reflect the position or policy
of the U.S.\ Government, and no official endorsement should be inferred.

\bibliographystyle{acl}
\bibliography{references}
\end{document}